\documentclass[twocolumn]{article}
\usepackage[a4paper, total={7in, 9.6in}]{geometry}

\usepackage{algpseudocode}
\usepackage{amssymb}
\usepackage{amsmath}
\usepackage{array}
\usepackage{float}
\usepackage{caption}
\usepackage{subcaption}
\usepackage{marginnote}
\usepackage{arydshln}

\usepackage{hyperref}
\hypersetup{
  colorlinks=true,
  linkcolor=black,
  citecolor=black,
  urlcolor=black
}

\newtheorem{definition}{Definition}
\newtheorem{example}{Example}
\newtheorem{theorem1}{Theorem}
\newtheorem{proof}{Proof}
\newtheorem{instantiation}{IPA instantiation}

\newcommand{\agentset}{\ensuremath{\mathcal{A}}}

  \usepackage{graphicx}
  \graphicspath{{./img/}{./performance_test_plot/IPM/}{./performance_test_plot/SeqIPM/}}  
  \DeclareGraphicsExtensions{.pdf,.jpeg,.png}

\begin{document}

\title{Data-driven Exploration of Mobility Interaction Patterns}

\newcommand{\alignauthor}{}
\newcommand{\affaddr}[1]{#1}
\newcommand{\email}[1]{#1}

\author{
\alignauthor
Gabriele Galatolo\\
       \affaddr{Dipartimento di Informatica}\\
       \affaddr{Universit\`{a} di Pisa, Italy}\\[2mm]
\alignauthor
Mirco Nanni\\
       \affaddr{KDDLab - ISTI}\\
       \affaddr{Consiglio Nazionale delle Ricerche, Italy}\\
       \email{mirco.nanni@isti.cnr.it}
}

\date{}

\maketitle

\begin{abstract}
Understanding the movement behaviours of individuals and the way they react to the external world is a key component of any problem that involves the modelling of human dynamics at a physical level.
In particular, it is crucial to capture the influence that the presence of an individual can have on the others. 
Important examples of applications include crowd simulation and emergency management, where the simulation of the mass of people passes through the simulation of the individuals, taking into consideration the others as part of the general context.
While existing solutions basically start from some preconceived behavioural model, in this work we propose an approach that starts directly from the data, adopting a data mining perspective.
Our method searches the mobility events in the data that might be possible evidences of mutual interactions between individuals, and on top of them looks for complex, persistent patterns and time evolving configurations of events.
The study of these patterns can provide new insights on the mechanics of mobility interactions between individuals, which can potentially help in improving existing simulation models.
We instantiate the general methodology on two real case studies, one on cars and one on pedestrians, and a full experimental evaluation is performed, both in terms of performances, parameter sensitivity and interpretation of sample results.
\end{abstract}

\mbox{}\\
{\bf Categories and Subject Descriptors:} H.2.8 Database
applications: Data mining\\
{\bf General Terms:} Algorithms

\section{Introduction}

{\bf Context.}
Modelling human movements in the physical world is a key ingredient in several important application contexts, most notably those involving efficient flows of masses (pedestrians, cars, etc.), risk management for crowded events and evacuation plans in case of emergencies.
Typical solutions for these applications are based on some behavioural model of single individuals within the mass of people, that determines the way he/she moves and how the movement is affected by the presence and the behaviour of the crowd around him/her.
Several models have been proposed in the last decade, that seem to capture various aspects of collective human movements in various contexts.
Some of them simplify the problem by viewing people as a cloud of particles that follow classical physical laws (e.g. \cite{helbing_panic_nature} on pedestrians); others follow a cognitive perspective, and see individuals as agents equipped with a set of decision rules (e.g. \cite{ngsim} for cars on highways); others, finally, try to integrate both aspects together~\cite{abm}.

{\bf Simulation models vs. data-driven patterns.}
The common element of the solutions mentioned above is that all the process starts from a predefined model, and assumes that it properly describes the reality.
Then, complying with the classical scientific method, the validity of the model is tested against some measurements of the world, such as ad hoc experiments or some dataset previously collected (often for other purposes).
Therefore, the accumulation of successful tests increase the confidence in the goodness of the model.

{\bf Our objectives.}
What we propose in this paper is basically to change perspective to the problem, and move a first step along the opposite direction.
The idea is to develop a bottom-up, data-driven approach to the understanding of human (physical) movements, that extracts directly from data the key events and patterns that appear consistently in the mobility of people.
A particular emphasis is given to the interaction among individuals, which is a fundamental component of human movement in mass mobility, such as crowds and road traffic.

{\bf Novel contributions.}
To summarize, the original contributions brought by this work are the following:
\begin{itemize}
\item first, a theoretical framework is provided that leads to the notion of basic interaction events and defines how they can be derived from raw mobility data (movement trajectories);
\item second, the problem of extracting significant interaction patterns (configurations of basic events) and evolution patterns is introduced;
\item third, the general framework is instantiated throughout the paper to a real case, later used in experiments;
\item fourth, a complete set of algorithms for extracting our patterns is presented, together with their computational complexity;
\item finally, a wide experimentation on two real datasets is carried out, one on cars and one on pedestrians, showing both performances and an exploration of the results obtained.
\end{itemize}

{\bf Paper organization.}
In the rest of the paper, we first introduce our complete framework for interaction pattern analysis (Section~\ref{sec:theory}); then we provide a set of pattern mining algorithms for our problem (Section~\ref{sec:algorithms}) and experiment them on real datasets (Section~\ref{sec:experiments}); finally, we discuss the relations with existing literature (Section~\ref{sec:related_works}) and conclude with some final remarks on future works (Section~\ref{sec:conclusion}). 

\section{Interaction Pattern Analysis}\label{sec:theory}

In this section we give the theoretical basis for the construction of patterns of interaction. 
We describe how to identify events between pairs of mobile agents, how to use them to build static interaction patterns, and how to extract evolution patterns.

\subsection{Basic Framework and Interaction Events} \label{sec:ipa}

In this section we introduce a general framework for the identification and analysis of events generated by single agents and by pairs of agents that interact. The \textit{Interaction Pattern Analysis} (IPA, from now on) framework is based on trajectory data. A {\em trajectory dataset} is defined as a set of $n$ trajectories $\mathbb{T} = \{\mathbb{T}_1, ... , \mathbb{T}_{n}\}$ observed in a {\em time scope} $T = \{ t \in \mathbb{N} | 0 \leq t \leq T_{max} \}$, and where $n$ is the number of unique agents of the dataset. Each trajectory of $\mathbb{T}$ is associated to a unique agent:

\begin{definition}[Agent set and agent]
\quad Given a set of trajectories $\mathbb{T}$ where $|\mathbb{T}| = n$, we define the agent set as
$\mathcal{A} = \{ \mathcal{A}_1, ... \mathcal{A}_n \}$, 
where a single agent is defined as
$
\mathcal{A}_{id} = \langle T_{id}, \{(\mathbb{T}^k_{id}, d^k_{id}, v^k_{id})\}_{k \in T_{id}} \rangle
$, with $1 \leq id \leq n$ and $T_{id} = [s,e] \subseteq T$. 
The values $\mathbb{T}^k_{id} = (x_k, y_k)$, $d^k_{id}$ and $v^k_{id}$ represent respectively the position of the agent $id$, its direction and its velocity in the instant $k$. For the sake of simplicity we can also refer to the set that represents an agent as
$\mathcal{A}_{id} = \{ \mathcal{A}^s_{id}, ... ,\mathcal{A}^e_{id}\}$, were, as above, $T_{id} = [s,e]$.
\end{definition}


The basic step in the analysis of interactions between agents consists in understanding which pairs of agents are actually close enough, in each time instant, to have the chance of interacting with each other.
This information is summarized in a mathematical structure, similar to temporal graphs~\cite{Kostakos}:

\begin{definition}[Neighbor Graph]
Given a set $\agentset$ of agents over the time scope $T$, and given a neighborhood function $\mathcal{N}$ defined over agents in a time instant, the {\em Neighbor graph} is a graph $G_{\agentset,T}^\mathcal{N} = \langle N, E \rangle$, with:

\begin{enumerate}
\item nodes $N = \{ \agentset_i^t ~|~ \agentset_i\in \agentset \wedge t \in T_i\}$
\item edges $E = E_{ego} \cup E_{int}$, where 
\[
E_{ego} = \{ (\agentset_i^t, \agentset_i^{t+1}) \in N^2 \}
\]
and
\[
E_{int} = \{ (\agentset_i^t, \agentset_j^t) \in N^2 ~|~  \agentset_j^t \in \mathcal{N}(\agentset_i^t) \wedge \agentset_i^{t-1}, \agentset_j^{t-1} \in N \}
\]
\end{enumerate}
\end{definition}

{\em Ego} edges represent vertical links that connect the different occurrences of each agent at consecutive time instants, whereas {\em interaction} edges represent pairs of agents that at specific moments fall within each other's sphere of influence, thus could potentially interact.
Notice that $E_{int}$ excludes initial time instants of agents: this is due to the fact that in our framework we will consider only events that involve changes of features between time instants, which are not defined for initial points. 

\begin{instantiation}[Neighborhood]
The neighborhood function can be instantiated in several ways, depending on the specific application context. As an example, in the experiments shown later in this paper we adopted a variant of visibility graphs: given a search diameter $d_{search}$, and describing agents by means of a center and a radius $r_{agent}$ (fixed for all agents and representing a radius of visibility), $\mathcal{N}$ returns all the agents visible to it, i.e. all the agents within distance $d_{search}$ for which we can draw a line that has no intersection with other circles between the two points.
\end{instantiation}

Our objective, now, is to enrich the neighbor graph with measures describing how agents change in time, and how they relate with each other.

\begin{definition}[Ego and interaction functions]
\mbox{}\\Given an agent set $\agentset$, a {\em ego function} $f^\agentset_{ego}$ is defined as a function that associates a fixed number $n_{ego}$ of real values to two consecutive instances of an agent, i.e. $f^\agentset_{ego}(\agentset_i^{t-1}, \agentset_i^{t})$ returns values in $\mathbb{R}^{n_{ego}}$.
$f^\agentset_{ego}$ is undefined for the first instance of an agent, i.e. ${t-1} \not\in T_i \Rightarrow f^\agentset_{ego}(\agentset_i^{t-1},\agentset_i^t)=undef$.

Similarly, a {\em interaction function} $f^\agentset_{int}$ associates a predefined number $n_{int}$ of real values to each pair of contemporary instances of two agents, also involving their previous instances, i.e. $f^\agentset_{int}(\agentset_i^{t-1},\agentset_i^t, \agentset_j^{t-1}, \agentset_j^t)$ returns values in $\mathbb{R}^{n_{int}}$.
As for $f^\agentset_{ego}$, $f^\agentset_{int}$ remains undefined when the first instance of an agent is involved.
\end{definition}

\begin{definition}[Labelled Graph]\label{def:labelled_graph}
Given a set $\agentset$ of agents over the time scope $T$, a neighborhood function $\mathcal{N}$, a ego function $f^\agentset_{ego}$ and a interaction function $f^\agentset_{int}$, the {\em Labelled graph} is a graph $G^L = \langle N, E, \mathcal{L_E} ,\mathcal{L_I} \rangle$, where:

\begin{enumerate}
\item $\langle N, E \rangle$ is the neighbor graph associated to $\agentset$ and $\mathcal{N}$
\item for each edge $(\agentset_i^{t-1}, \agentset_i^{t}) \in E$, a label $\mathcal{L_E}(\agentset_i^{t})$ is associated to node $\agentset_i^{t}$, with $\mathcal{L_E}(\agentset_i^{t}) = f^\agentset_{ego}(\agentset_i^{t-1},\agentset_i^t)$
\item for each edge $(\agentset_i^{t}, \agentset_j^{t}) \in E$, a label $\mathcal{L_I}(\agentset_i^{t},\agentset_j^{t}) = f^\agentset_{int}(\agentset_i^{t-1},\agentset_i^t,\agentset_j^{t-1},\agentset_j^t)$ is associated to edge $(\agentset_i^{t}, \agentset_j^{t})$.

\end{enumerate}
\end{definition}

We briefly describe an instantiation of the interaction functions that we will use in the rest of the paper. 

\begin{instantiation}[Interaction functions]
\mbox{}\\
The set of interaction functions adopted in this work is summarized in table \ref{tab:interactions_functions}. 
In the {\em distance} function a negative value indicates an (instantaneous) approach between agents, while a positive value indicates a distancing. In the {\em velocity} function we can derive by the returned value if the first agent is faster or slower than the second one. The {\em direction} function calculates the difference of direction between the two agents, ranging between 0 (same direction) and 180 (opposite direction). The {\em position} function computes the position of the first agent w.r.t. the second one and it returns a code that indicates if the agent $\agentset_i$ is behind ($v_1$), ahead of ($v_2$), lateral to ($v_3$), flanked to with same direction ($v_4$), moving in front of ($v_5$), moving behind the other in opposite direction ($v_6$) or not moving as like agent $\agentset_j$ ($v_7$): the function analyzes the line generated by the two points pair $(\mathbb{T}_i^{t-1}, \mathbb{T}_i^t)$ and $(\mathbb{T}_j^{t-1}, \mathbb{T}_j^t)$, combining the information about the degree of parallelism of the two lines, the lateral positioning of the agent and the moving indication to return the correct positioning. 
The last function ({\em align}) checks the alignment between two agents: indicating with $r_i$ the line that passes through $\mathbb{T}_i^{t-1}$ and $\mathbb{T}_i^{t}$, the function checks if the second agent, in two consecutive instants, is contained into a buffer of radius $\epsilon_{align}$ drawn around $r_i$. 
In the framework instantiation considered here, no ego functions are used.

\begin{table} 
\renewcommand{\arraystretch}{1.3} 
\caption{Interaction functions used in the paper.}
\label{tab:interactions_functions} 
\centering 
\resizebox{\linewidth}{!}{%
\begin{tabular}{c || c} 
\hline 
\bfseries name & \bfseries FUNCTION DEFINITION \\ 
\hline\hline 
distance  & $|| \mathbb{T}_i^t - \mathbb{T}_j^t ||_2 - || \mathbb{T}_i^{t-1} - \mathbb{T}_j^{t-1} ||_2$ \\ 
\hline 
velocity & $v^t_i - v^t_j$ \\
\hline
direction & $min\{ 360 - |d^t_i - d^t_j|, |d^t_i - d^t_j|\}$\\
\hline
position & $position(\agentset_i^{t-1},\agentset_i^{t},\agentset_j^{t-1}, \agentset_j^{t}, \epsilon_{lat}, \epsilon_\parallel, \epsilon_{move})$ \hfill $\rightarrow$ see text\\
\hline
align & $\begin{cases}1 & \mbox{if } d(r_i, \mathbb{T}_j^{t-1}) \leq \epsilon_{align} \wedge d(r_i,\mathbb{T}_j^t)\leq \epsilon_{align}\\ 0 & otherwise \end{cases}$ \\
\hline
\end{tabular}}
\end{table} 
\end{instantiation}

The functions introduced above are building blocks that can be combined to define events of several kinds.

\begin{definition}[Event template]\label{def:event_templates}
Given an agent set \agentset{} and a time scope $T$, an {\em event template} is defined as a predicate $P: \agentset \times \agentset \times T^2 \rightarrow \mathcal{B}$.
The intended meaning of $P(\agentset_i, \agentset_j, a, b)$ is that property $P$ holds in time interval $[a,b]$, involving agents $\agentset_i$ and $\agentset_j$.
\end{definition}

This definition includes not only properties that hold on each single time instant, separately from the others, but also cases that depend on what happens collectively throughout interval $[a,b]$.

\begin{example}
An example of event template is the following: $\mathit{approach}(\agentset_1, \agentset_2, t_s, t_e)$ which holds iff for each time instant $t \in [t_s,t_e)$ is satisfied $f_{distance}(\agentset^t_1, \agentset^t_2) < \epsilon_{dist}$ or $f_{distance}(\agentset^t_1, \agentset^t_2) \geq  \epsilon_{dist} \Rightarrow f_{distance}(\agentset^{t+1}_1, \agentset^{t+1}_2) <  \epsilon_{dist}$.
\end{example}

This example shows a case of complex form, not limited to a simple monotonic behaviour. Indeed, non-strict decreases in the distance are allowed, but only if isolated. This kind of behaviour was empirically observed to be rather frequent in the datasets we considered in the experiments of this paper. For this reason, we provide here a formal definition of such class of events, and it will later be adopted for the event types considered in this work:

\begin{definition}[Quasi-continuous events]
An event of form {\em P$(\agentset_i,\agentset_j, t_s, t_e)$} is said to be {\em quasi-continuous} iff for each $t \in [t_s, t_e]$ the property that defines $P$ holds or at least it holds at time $t+1$ (option possible only if $t < t_e$).
\end{definition}

\begin{table*} 
\renewcommand{\arraystretch}{1.3} 
\caption{Definition of events used in the paper.} 
\label{concrete_events} 
\centering 
\resizebox{\linewidth}{!}{%
\begin{tabular}{c || c || c } 
\hline 
\bfseries Legend & \bfseries CONDITION & \bfseries SUBJECT/OBJECT \\ 
\hline\hline 
moving\_away & $f_{distance}(\agentset_i,\agentset_j) > \epsilon_{dist}$ & $\agentset_i$ is the subject if it's faster than $\agentset_j$ \\
\includegraphics[scale=0.5]{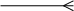}& & \\
\hline 
following & $f_{direction}(\agentset_i,\agentset_j) \leq \epsilon_{direction} \wedge f_{aligned}(\agentset_i,\agentset_j) = 1 \wedge$ & $v$ indicate that $\agentset_i$ is behind the agent $\agentset_j$, $\agentset_i$ it's the subject if it follows the $\agentset_j$ \\
\includegraphics[scale=0.5]{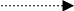}& $f_{position}(\agentset_i,\agentset_j) = v \wedge -\epsilon_{dist} \leq f_{distance}(\agentset_i,\agentset_j) \leq \epsilon_{dist}$ & \\
\hline
maintaining\_distance & $-\epsilon_{dist} \leq f_{distance}(\agentset_i,\agentset_j) \leq \epsilon_{dist}$ & $\agentset_i$ is the subject if it follows $\agentset_j$ \\
\includegraphics[scale=0.5]{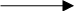}&$\wedge \nexists [t_{s'}, t_{e'}) \subseteq [t_s, t_e) \, . \, t_{e'} - t_{s'} > 1 \wedge \,${\em following}$(\agentset_i,\agentset_j, t_s, t_e)$& \\
\hline
back\_aligned\_approach & $f_{direction}(\agentset_i,\agentset_j) \leq \epsilon_{direction} \wedge f_{aligned}(\agentset_i,\agentset_j) = 1 \wedge$ & $v$ indicate that $\agentset_i$ is behind the agent $\agentset_j$, $\agentset_i$ it's the subject if it follows the $\agentset_j$ \\
\includegraphics[scale=0.5]{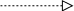} & $f_{position}(\agentset_i,\agentset_j) = v \wedge f_{distance}(\agentset_i,\agentset_j) < -\epsilon_{dist}$& \\
\hline
frontal\_approach & $f_{direction}(\agentset_i,\agentset_j) \geq (180 - \epsilon_{direction}) \wedge f_{aligned}(\agentset_i,\agentset_j) = 1 \wedge$ & $v$ indicate that $\agentset_i$ and $\agentset_j$ are opposite, $\agentset_i$ it's the subject if it's faster than $\agentset_j$ \\
\includegraphics[scale=0.5]{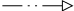}&$f_{position}(\agentset_i,\agentset_j) = v \wedge f_{distance}(\agentset_i,\agentset_j) < -\epsilon_{dist}$& \\
\hline
opposite\_approach & $f_{direction}(\agentset_i,\agentset_j) \geq (180 - \epsilon_{direction}) \wedge f_{aligned}(\agentset_i,\agentset_j) = 0 \wedge$ & $v$ indicate that $\agentset_i$ and $\agentset_j$ are opposite, $\agentset_i$ it's the subject if it's faster than $\agentset_j$ \\
\includegraphics[scale=0.5]{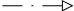}&$f_{position}(\agentset_i,\agentset_j) = v \wedge f_{distance}(\agentset_i,\agentset_j) < -\epsilon_{dist}$& \\
\hline
approach & $f_{distance}(\agentset_i,\agentset_j) < -\epsilon_{dist} \wedge \nexists [t_{s'}, t_{e'}) \subseteq [t_s, t_e) \, . \, t_{e'} - t_{s'} > 1 \wedge$ & $\agentset_i$ it's the subject if it's faster than $\agentset_j$ \\
\includegraphics[scale=0.5]{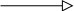} & ({\em back\_aligned\_approach$(\agentset_i,\agentset_j, t_s, t_e)$} &\\
& $ \vee ${\em frontal\_approach$(\agentset_i,\agentset_j, t_s, t_e)$}$ \vee ${\em opposite\_approach$(\agentset_i,\agentset_j, t_s, t_e)$})& \\
\hline
flanking & $f_{direction}(\agentset_i,\agentset_j) \leq \epsilon_{direction} \wedge f_{position}(\agentset_i,\agentset_j) = v$ & $v$ indicate that $\agentset_i$ is flanking the agent $\agentset_j$, $\agentset_i$ it's the subject if it's faster than $\agentset_j$  \\
\includegraphics[scale=0.5]{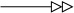}& & \\
\hline 
opposite\_flanking & $f_{direction}(\agentset_i,\agentset_j) \leq (180 - \epsilon_{direction}) \wedge f_{position}(\agentset_i,\agentset_j) = v$ & $v$ indicate that $\agentset_i$ is flanking the agent $\agentset_j$, $\agentset_i$ it's the subject if it's faster than $\agentset_j$  \\
\includegraphics[scale=0.5]{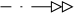}& & \\
\hline 

\end{tabular}}
\end{table*}

\begin{instantiation}[Event templates]
The event templates adopted in the experiments of this paper are reported in table~\ref{concrete_events}.
For the sake of conciseness, the functions of form $f_{name}(\agentset_i^{t-1},\agentset_i^{t},\agentset_j^{t-1},\agentset_j^{t})$ will be represented by $f_{name}(\agentset_i,\agentset_j)$.
On the left column, the table also contains the legend of the symbols that we will use to identify the events between agents in figures in the rest of the work.
\end{instantiation}

After we defined which kind of events we are interested in, we can finally look for their occurrences in our data, exploiting the concise graph representation given in Definition~\ref{def:labelled_graph}.

\begin{definition}[Interaction Events]
\quad Given a labelled graph $G^L = \langle N, E, \mathcal{L_E} ,\mathcal{L_I} \rangle$ and a family $\mathcal{ET}$ of event templates, the {\em graph events} of $G^L$ are defined as the set $\mathcal{EI}$ of all event instances $P(\agentset_i, \agentset_j, a, b)$ such that:
\begin{enumerate}
\item $\forall t\in [a,b]. \agentset_i^t, \agentset_j^t \in N$
\item $P(\agentset_i, \agentset_j, a, b) = \mathit{true}$
\item $[a,b]$ is maximal, i.e.: $[c,d] \supset [a,b] \Rightarrow \neg P(\agentset_i, \agentset_j, c, d)$
\end{enumerate}

\end{definition}

\subsection{Static Interaction Patterns}
\begin{figure}[h]
  	\centering
    \includegraphics[scale=0.4]{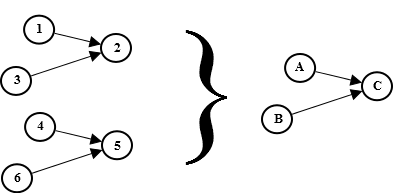} \\
	Static pattern instances \hspace*{10mm}Static pattern \qquad\qquad
	\caption{Left: instances of a complex event involving different agents. Right: pattern representing the two instances.}
	\label{fig:first_example_ip_gp}
\end{figure}

The basic IPA framework presented in the previous section allows to identify all the occurrences of single events that follow the templates chosen by the end-user.
Beside providing a first level of information that can be useful for some applications, they can also be the starting point for locating more complex schemes hidden in the data.
In this section we introduce the notion of {\em static interaction patterns}, that capture groups of events that appear frequently together. For instance, we can expect that in the context of cars moving on a highway it will be easy to find groups of agents that maintain their mutual distance almost constant for a significant time. 
Figure~\ref{fig:first_example_ip_gp} on the left, shows two examples of patterns, each involving two agents that maintain their distance from a third one.
When looking for a pattern, we are interested in finding a general {\em schema} that represents the different yet equivalent single instances.
For example, the two instances on the left of Figure~\ref{fig:first_example_ip_gp} can be concisely represented by the pattern on the right, where specific agent IDs (from 1 to 6) are replaced with variables (from A to C).

We first need to define what kind of pattern instances we are interested in.
Our choice is to require that a pattern instance persists for at least a minimum duration, to avoid spurious events. Also, we require that for a given set of agents involved the instance is maximal, in the sense of including as many events as possible relative to that set of agents, provided that the persistence requirement mentioned above is satisfied.

\begin{definition}[Static Pattern Instance]\label{definition:instance_pattern}
Given a \mbox{} minimum interval length $t_{min}$, a set of event instances $pi = \{e_1, ..., e_{k}\} \subseteq \mathcal{EI}$ is said to be a {\em static pattern instance} if the following properties are satisfied:

\begin{enumerate}
\item (Persistence) $\displaystyle I_{pi}=\bigcap^{k}_{i = 1} interval(e_i) \wedge \left|I_{pi}\right| \geq t_{min}$;

\item (Connectedness) $\forall e_i, e_j\in pi, \exists \{e^1, \ldots, e^m\}\subseteq pi. e^1=e_i \wedge e^m=e_j\wedge \forall_{1\leq l < m}.  agents(e^l) \cap agents(e^{l+1}) \neq \emptyset$;

\item (Maximality) $\displaystyle e\in \mathcal{EI} \wedge agents(e)\subseteq \bigcup^k_{i=1} agents(e_i) \wedge |I_{pi} \cap interval(e)| \geq t_{min} \Rightarrow e\in pi$;

\end{enumerate}

where $interval(e) = [a,b]$ and $agents(e)=\{\agentset_i,\agentset_j\}$ for $e=P(\agentset_i,\agentset_j,a,b)$.
\end{definition}

The {\em connectedness} condition, in particular, states that the multi-graph formed by a static pattern instance -- where the nodes are the agents involved and each event in the pattern corresponds to an edge between two agents -- should be connected.
That ensures that all the events in each pattern are somehow related to each other.

The next step is to group together all those pattern instances that are actually instances of the same pattern.
The basic idea is that such instances share the same structure, only differing for the identity of the agents involved.
Grouping pattern instances, therefore, can be seen as finding a renaming of agents that makes all instances identical.
We formalize this in terms of isomorphism among instances and equivalence classes.

\begin{definition}[Isomorphic Instances]
\quad Given two  static pattern instances $pi_1$ and $pi_2$, we say that they are {\em isomorphic instances}, denoted with $pi_1 \simeq pi_2$, if there exists a bijective function $\phi: \agentset \rightarrow \agentset$ such that: $\forall \agentset_i, \agentset_j \in \agentset. \forall a,b,a',b' \in T. P(\agentset_i, \agentset_j, a, b)\in pi_1 \Leftrightarrow  P(\phi(\agentset_i), \phi(\agentset_j),$ $a', b')\in pi_2$. 
We will denote this fact also as $pi_1 \simeq_\phi pi_2$, to emphasize the specific mapping function $\phi$ chosen. 
\end{definition}



The isomorphism between instances corresponds to the traditional isomorphism among graphs.
In particular, our pattern instances are multi-graphs (different events can involve the same two agents) where the edges (events) are associated to a type (the kind of event) while nodes (the agents) are unlabelled, since we are not interested in the agent identity.

It is easy to see that the instance isomorphism is an equivalence relation, and therefore induces equivalence classes that can be used as representative patterns of a set of (isomorphic) instances.
In the following we exploit this to give a definition of frequent static patterns.

\begin{definition}[Frequent Static Pattern]\label{def:static_patterns}
Given \mbox{} the set $\mathcal{PI}$ of static pattern instances inferred from all event instances $\mathcal{EI}$, we define the set of {\em static patterns} in $\mathcal{PI}$ as $\mathcal{SP} = \{ [pi]_\simeq \, | \, pi \in \mathcal{PI}\}$, where $[pi]_\simeq$ denotes the equivalence class induced by $\simeq$ that contains $pi$. 
We define the {\em temporal support} of static pattern $sp \in \mathcal{SP}$ as 
$$
supp(sp) = \frac{\left|  \bigcup_{pi\in \mathcal{PI}\wedge [pi]_\simeq=sp} interval(pi) \right|}{|T|}.
$$
A static pattern $sp$ is said to be {\em frequent} iff $supp(sp) \geq t_\mathit{supp}$, where $t_\mathit{supp} \in [0,1]$ is a minimum temporal support threshold.
Finally, we denote with $\mathcal{FSP}$ the set of all frequent static patterns in $\mathcal{SP}$.
\end{definition}

%

Our choice for a support threshold based on the temporal coverage of patterns, rather than the number of pattern instances, is motivated by the fact that we want to give emphasis to persistence and pervasiveness of patterns.
Counting the number of instances cannot do that in our case, since a frequent pattern might be the result of a burst of instances within a short time, which might be meaningful for identifying exceptional phenomena, but much less useful for detecting common behaviours.

When useful for visualization and discussion purposes, in the rest of the paper we will adopt the convention already used Figure~\ref{fig:first_example_ip_gp}, where equivalence classes $[pi]_\simeq$ are represented as distinct capital letters.

\subsection{Evolving Patterns} \label{sec:evolving_patterns_sec}
\begin{figure}[h]
  	\centering
    \includegraphics[scale=0.45]{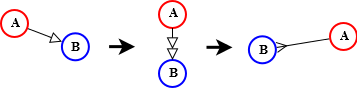}
	\caption{A complete overtaking: an \textit{approach}(A,B) followed by a \textit{flanking}(A,B) and finally by a \textit{moving\_away}(A,B)}
	\label{fig:complete_overtaking}
\end{figure}

In the previous section we have introduced the notion of interaction pattern, which abstract from the identities of an instance's agents to describe a set of general complex schemes of events happening frequently together.
Once discovered these complex schemes we could ask if there is some consequentiality between the static patterns found, and in which form it appears. For example, analyzing a dataset involving moving cars, one can expect to find sequences of patterns that represent an overtaking, which might have the form shown in figure \ref{fig:complete_overtaking}.

We formalize the problem through the following definitions, in a manner similar to static patterns.

\begin{definition}[Sequence pattern instances] \label{def:sequence_instance}
\mbox{}\\
Given a set of frequent static patterns $\mathcal{FSP}$, a temporal window $t_{win} \in \mathbb{N}$ and a $min_{jc} \in [0,1]$, a sequence of instances $spi = (pi_1, pi_2, \ldots pi_k)$, where $\forall 1 \leq j \leq k \, . \, [pi_j]_{\simeq} \in \mathcal{FSP}$, is said to be a {\em sequence pattern instance} if the following conditions are verified:

\begin{enumerate}

\item $jc({pi_1},{pi_2}) \geq min_{jc} \,\vee\, agents({pi_1}) \subseteq agents({pi_2}) \vee agents({pi_2}) \subseteq agents({pi_1})$.

\item $\forall 1 < i < k: jc(pi_i, pi_{i+1}) \geq min_{jc}$

\item $\forall 1 \leq i < k, t^i_{start} < t^{i+1}_{start}$

\item $(t^k_{start} - t^1_{start}) \leq t_{win}$

\item $\forall 1 \leq i \leq k: kernel(spi) \subseteq agents(pi_i)$

\end{enumerate}

where $jc(a,b)$ denotes the Jaccard coefficient between the sets $agents(a)$ and $agents(b)$, and for each $pi_i\in spi$ we have $interval(pi_i)=[t^i_{start},t^i_{end}]$.
The set of agents $kernel(spi) = agents({pi_1}) \cap agents({pi_2})$ is called the {\em kernel} of the sequence, while the {\em interval of the sequence} $(pi_1, \ldots, pi_k)$ is defined as $interval(pi_1, \ldots, pi_k) = \bigcup^k_{i = 1 ... k} interval(pi_i)$. 
The set of all the sequence pattern instances is denoted with $\mathcal{SPI}$.
\end{definition}

We observe that in this definition we give emphasis to the first two elements of the sequence, since they define the core set of agents to be tracked along the sequence, though a limited amount of agents can leave or join it at each step of the sequence ($jc(pi_i, pi_{i+1}) \geq min_{jc}$). 

As for the case of static patterns, several sequence instances are actually instances of the same general pattern. Since the latter represents our objective, we use again equivalence classes of instances to represent general patterns.

\begin{definition}[Isomorphic sequence instances]
\mbox{}\\
Given two sequence pattern instances $spi_1 = (pi_1^1, \ldots, pi_1^n)$ and $spi_2 = (pi_2^1, \ldots, pi_2^m)$, we say that they are {\em isomorphic sequence instances}, denoted with $spi_1 \simeq spi_2$, if (i) $n=m$, and (ii) there exists a bijective function $\phi: \agentset \rightarrow \agentset$ such that $\forall 1\leq i \leq n. pi_1^i \simeq_\phi pi_2^i$.
\end{definition}

In the definition above, we notice that the name mapping function $\phi$ is the same for all static pattern instances involved, meaning that if an agent appears in more than one static instance within a sequence, the {\em new identities} it will have after the name mapping will be the same across the whole sequence.

Isomorphic sequences will be the basis for defining the notion of evolving interaction patterns.

\begin{definition}[Evolving interaction patterns]
\mbox{}\\
Given the set $\mathcal{SPI}$ of all sequence pattern instances, we define the set of {\em evolving interaction patterns} in $\mathcal{SPI}$ as the set $\mathcal{EP} = \{ [spi]_{\simeq} \: | \: spi \in \mathcal{SPI}\}$ where $[spi]_{\simeq}$ denotes the equivalence class induced by $\simeq$ that contains $spi$. 
We define the {\em temporal support} of an evolving interaction pattern $ep \in \mathcal{EP}$ as 
$$
supp(ep) = \frac{\left|  \bigcup_{spi\in \mathcal{SPI}\wedge [spi]_{\simeq}=ep} interval(spi) \right|}{|T|}.
$$
An evolving interaction pattern $ep$ is said to be {\em frequent} iff $supp(ep) \geq t_\mathit{e\!-\!supp}$, where $t_\mathit{e\!-\!supp} \in [0,1]$ is the minimum temporal support threshold for evolving patterns.
Finally, we denote the set of all frequent evolving patterns with $\mathcal{FEP}$.
\end{definition}


\section{Pattern Mining Algorithms}\label{sec:algorithms}

In this section we provide the algorithms we developed to find the frequent static and evolving patterns. 
The implementation of the IPA framework is relatively straightforward and computationally inexpensive, at least for the instantiation considered in this paper, therefore, for the sake of presentation clarity we omit its description here. Details can be found in~\cite{tesi_gabriele}.

\subsection{Computing Frequent Static Patterns}
\label{sec:computing_FSP}
To compute the frequent static patterns we will start generating all the static pattern instances from the set $\mathcal{EI}$ of events found from the IPA framework, following Definition~\ref{definition:instance_pattern}. 
A first important task for computing frequent static patterns is the detection of isomorphisms among pattern instances. 
As we mentioned in the previous section, this task can be rephrased as a graph isomorphism problem, which is known to be NP-Hard.
In order to make it tractable in practise, we developed an heuristic to reduce considerably the number of comparisons necessary.
We base our strategy on two main considerations: first, since we work with labelled multi-graphs, if two vertices have the same number of outgoing edges but have a different number of edges with the same label, the two graphs cannot be isomorphic; second, looking at the adjacency matrices of two graphs representing pattern instances, only the vertices that have the same number of outgoing edges with the same labels and that have the same incoming degree can represent a possible mapping of two vertices. 
Then we can use this fact to generate only the necessary subset of permutations of the adjacency matrix of one of the two matrices to know if the two graphs are isomorphic.



We are now ready to present the overall \textit{SIPM} (Static Interaction Pattern Mining) algorithm in Figure~\ref{alg:SIPM_code}.
The algorithm follows a classical level-wise generation of candidates {\em a la} Apriori, exploiting a straightforward monotonicity property: the more agents we add to a form a pattern, the shorter is their collective persistence, i.e. the smaller the temporal support of the pattern.
At line \ref{lst:extract2ip} we extract from $\mathcal{EI}$ all the pairs of vertices that form a pattern instance as formalized in Definition~\ref{definition:instance_pattern};
the main cycle of the algorithm starts on line \ref{lst:main_cycle} and at each step $k$ builds all the patterns that involve $k+1$ distinct agents, starting from those of size $k$ found in the previous iteration;
 at line \ref{lst:merge_group} we group all the pattern instances that are isomorphic to form equivalence classes (i.e., static patterns). 
The cycle starting from \ref{lst:persistence_groups} is the one that recognizes a static pattern as frequent if the set of its instances covers a sufficiently large portion of time. 
A key step of the algorithm is the generation of the candidates (line \ref{lst:candidate_generation}), which is described in detail in Figure~\ref{alg:candidate_generation_code}.

\begin{figure}[h]
	\begin{algorithmic}[1]
	{\scriptsize
	\Procedure{SIPM}{$\mathcal{EI}, t_\mathit{supp}, t_{min}$}		
		\State $k = 2$
		\State $\mathcal{FSP} = \emptyset$
		\State $C_k = extract\_2\_ip(\mathcal{EI}, t_{min})$ \label{lst:extract2ip}	
		\While{$C_k \neq \emptyset$} \label{lst:main_cycle}
			\State $I_k = \emptyset$
			\State $ \mathcal{SP}_k = merge\_isomorphic\_instances(C_k) $ \label{lst:merge_group}
			\ForAll{$sp \in \mathcal{SP}_k$} \label{lst:persistence_groups}
				\If{$supp(sp) \geq t_\mathit{supp} $}
					\State $I_k = I_k \cup (\bigcup_{[ip]_{\simeq} = sp} ip)$
					\State $\mathcal{FSP} = \mathcal{FSP} \cup \{ sp \}$
				\EndIf
			\EndFor
			\State $C_{k+1} = candidate\_generation(I_k, t_{min}) $ \label{lst:candidate_generation}
			\State $ k = k + 1; $
		\EndWhile\label{euclidendwhile}
		\State \textbf{return} $\mathcal{FSP}$
	\EndProcedure
	}
\end{algorithmic}
\caption{Static Interaction Pattern Mining pseudo-code}
\label{alg:SIPM_code}
\end{figure}

In the candidates generation we use a set of intermediate results to collect the interaction patterns of $k$ agents that are persistent, and we try to merge them to create new candidates containing $k+1$ agents. 
Since the instance patterns found this way could be not maximal, we perform a second recombination, on line \ref{lst-cg:remove_ir}, where we merge the intermediate results if they are persistent.
In the positive cases, we keep the maximal pattern found and remove its {\em generators} from the set of the intermediate results.

\begin{figure}
	\begin{algorithmic}[1]
	{\scriptsize
	\Procedure{candidate\_generation}{$I_k, t_{min}$}		
		\State $C = \emptyset$
		\ForAll{$ip_i \in I_k$}
		\State $IR = \emptyset, CR = \emptyset$ \label{lst-cg:ir}
			\State $IC = \{ip_j \in I_k |\; agents(ip_j) = \{agents(ip_i) \setminus \{id_b\}\} \cup id_c, \{id_a,id_b\} \subseteq ip_i, id_c \in \mathcal{N}(id_a) \setminus agents(ip_i) \}$
			\ForAll {$ip_j \in IC$}
				\If{$interval(ip_i) \cap interval(ip_j) \geq t_{min}$}
					\State $IR = IR \cup \{merge(ip_i, ip_j)\}$
					\State $CR = CR \cup \{merge(ip_i, ip_j)\},$ \label{lst-cg:merge}
				\EndIf
			\EndFor
			\ForAll {$ir_i, ir_j \in IR$ s.t. $agents(ir_i) = agents(ir_j)$}
				\If{$| interval(ir_i) \cap interval(ir_j) | \geq t_{min}$}
					\State $CR = \{CR \cup \{merge(ir_i, ir_j)\}\} \mathcal{n} \{ \{ir_i\} \cup \{ir_j\}\}$ \label{lst-cg:remove_ir}
				\EndIf
			\EndFor
			\State $C = C \cup CR$
		\EndFor
		\State \textbf{return} $C$
	\EndProcedure
	}
\end{algorithmic}
\caption{Candidate Generation pseudo-code}
\label{alg:candidate_generation_code}
\end{figure}


Analyzing the pseudo-code shown in Figure~\ref{alg:SIPM_code}, we can characterize the complexity of the algorithm as follows.

\begin{theorem1}[SIPM complexty]
Given a set of event instances $\mathcal{EI}$ observed in a time scope {\em T} and a minimum persistence time for each instance pattern of $t_{min}$, the time complexity of the SIPM algorithm is $O(m \cdot (4n)^n \cdot n^3)$, where $n$ is the maximum number of different agents present in one instant and $m = \lfloor \frac{|T|}{t_{min}} \rfloor$.
\end{theorem1}
\begin{proof}
Omitted. See~\cite{tesi_gabriele} for details.
\end{proof}

The complexity given by the previous theorem provides an upper bound that in the real cases is usually never reached. Indeed, the heuristics for isomorphism testing in union with the minimum temporal support constraint helps to reduce significantly the factor $(4n)^n$ in the complexity.

\subsection{The EvIPM algorithm}
Now we describe the \textit{evolving interaction patterns mining} algorithm (EvIPM), shown in figure \ref{alg:EvIPM_code}, which computes evolving patterns starting from the static patterns $\mathcal{FSP}$ extracted by SIPM and their corresponding instances in $\mathcal{PI}$.
Also in this case, the structure of the algorithm follows a pattern-growing schema in the style of Apriori.

\begin{figure}[t]
	\begin{algorithmic}[1]
	{\scriptsize
	\Procedure{EvIPM}{$\mathcal{FSP}, \mathcal{PI}, t_{win}, min_{jc}, t_\mathit{e\!-\!supp}$}		
		\State $k=2$
		\State $\mathcal{SPI}^{*}_2 = \{ (pi_1, pi_2)\in \mathcal{PI}^2 \: | \: 
			[pi_1]_\simeq \in \mathcal{FSP} 
			\wedge [pi_2]_\simeq \in \mathcal{FSP} 
			\wedge interval(pi_1) = [t^1_s, t^1_e]
			\wedge interval(pi_2) = [t^2_s, t^2_e]
			\wedge t^1_s < t^2_s \leq t^1_s + t_{win}
			\wedge ( agents(pi_1)\subseteq agents(pi_2) 
				\vee agents(pi_2)\subseteq agents(pi_1)
				\vee jc(pi_1,pi_2) \geq min_{jc})
			\}$ \label{lst-seq:sip2}
		\State $\mathcal{FEP} = \{ [spi]_\simeq \: | \: spi \in \mathcal{SPI}_2 \wedge supp([spi]_\simeq) \geq t_\mathit{e\!-\!supp} \}$ \label{lst-seq:fep2}
		\State $\mathcal{SPI}_2 = \{ spi \in \mathcal{SPI}^*_2 \:|\: [spi]_\simeq \in \mathcal{FEP} \}$\label{lst-seq:sip_fep2}
		\While{$\mathcal{SPI}_k \neq \emptyset$} \label{lst-seq:main_cycle_start}
			\ForAll{$spi = (pi_1, ..., pi_k) \in \mathcal{SPI}_k$}
				\State $[t_s, t_e] = interval(pi_1)$ \label{lst-seq:main_construct_start}
				\State $\mathcal{PI}_{tail} = \{pi \, | \, interval(pi)=[t_1,t_2] \wedge t_1\in[t_s,t_s+t_{win}] \wedge kernel(spi) \subseteq agents(ip) \wedge jc(pi_k, pi) \geq min_{jc} \}$ \label{lst-seq:tail}
				\State $\mathcal{SPI}_{k+1}^{spi} = \{ (pi_1, \ldots, pi_k, pi) | pi \in \mathcal{PI}_{tail}\}$\label{lst-seq:sipk}
			\EndFor
			\State $\mathcal{SPI}^*_{k+1} = \bigcup \mathcal{SPI}_{k+1}^{spi}$\label{lst-seq:sipk_all}
			\State $\mathcal{FEP} = \mathcal{FEP} \cup \{ [spi]_\simeq \:|\: spi\in \mathcal{SPI}_{k+1} \wedge supp([spi]_\simeq) \geq t_\mathit{e\!-\!supp} \}$  \label{lst-seq:frequent_set}
			\State $\mathcal{SPI}_{k+1} = \{ spi \in \mathcal{SPI}^*_{k+1} \:|\: [spi]_\simeq \in \mathcal{FEP} \}$ \label{lst-seq:frequent_set_spi}
			\State $k = k+1$
		\EndWhile \label{lst-seq:main_cycle_end}
		\State \textbf{return} $\mathcal{FEP}$
	\EndProcedure
	}
\end{algorithmic}
\caption{Evolving Interaction Pattern Mining pseudo-code}
\label{alg:EvIPM_code}
\end{figure}


The algorithm starts with the creation of sequence instances composed by two static pattern instances (step \ref{lst-seq:sip2}), by following the special case conditions described in Definition~\ref{def:sequence_instance}; then, we keep only those forming frequent evolving patterns (lines \ref{lst-seq:fep2}-\ref{lst-seq:sip_fep2}). The main cycle in lines \ref{lst-seq:main_cycle_start}-\ref{lst-seq:main_cycle_end} performs the same operations for the general case, trying to append a new static pattern (instance) to existing sequences at each step: in line \ref{lst-seq:tail} we find all the static pattern instances with starting time into the temporal window, and then we construct all the possible sequences of size $k+1$, creating a new sequence that is composed by the current $spi$ and one of the possible patterns found in $\ref{lst-seq:tail}$.
Again we keep only those sequences belonging to frequent evolving patterns, lines \ref{lst-seq:frequent_set}-\ref{lst-seq:frequent_set_spi}.

\begin{theorem1}[EvIPM complexty]
Let $n$ be the number of instances relative to all patterns contained in $\mathcal{FSP}$, $t_{win}$ the dimension of the temporal window, and $m$ the maximum number of instance patterns that start in the same instant $i \in T$. Then, the computational complexity of the EvIPM algorithm is given by $O(n \cdot m^t)$.
\end{theorem1}
\begin{proof}
Omitted. See~\cite{tesi_gabriele} for details.
\end{proof}

\section{Experiments}\label{sec:experiments}

In this section we summarize the results of a set of experiments performed with our framework and algorithms over two datasets. 
The system employed is an Intel i7-3517U 1.9/2.4 Ghz with 8GB of main memory, and our algorithms are all written in Java.

\begin{figure*}[t]
        \centering
        \begin{subfigure}[b]{0.24\textwidth}
            \includegraphics[width=\textwidth]{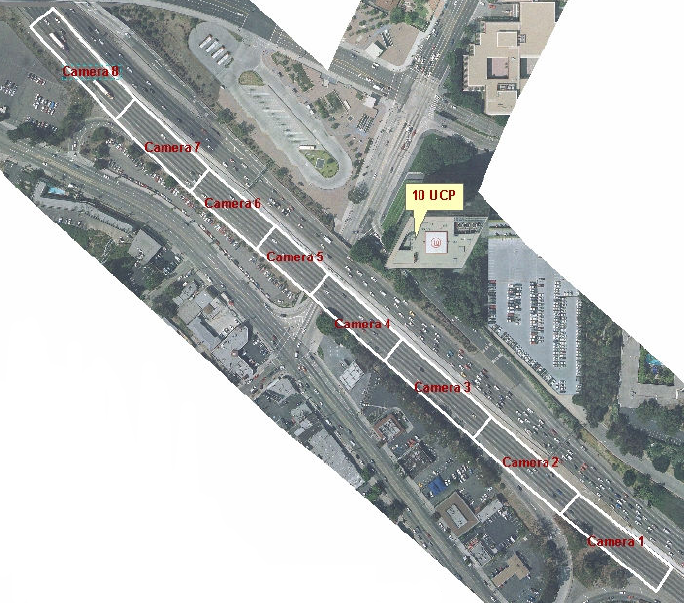}
            \caption{Area of NGSIM dataset}
    		\label{fig:ngsim_area}
        \end{subfigure}%
        \qquad
        \begin{subfigure}[b]{0.24\textwidth}
        	\includegraphics[width=\textwidth]{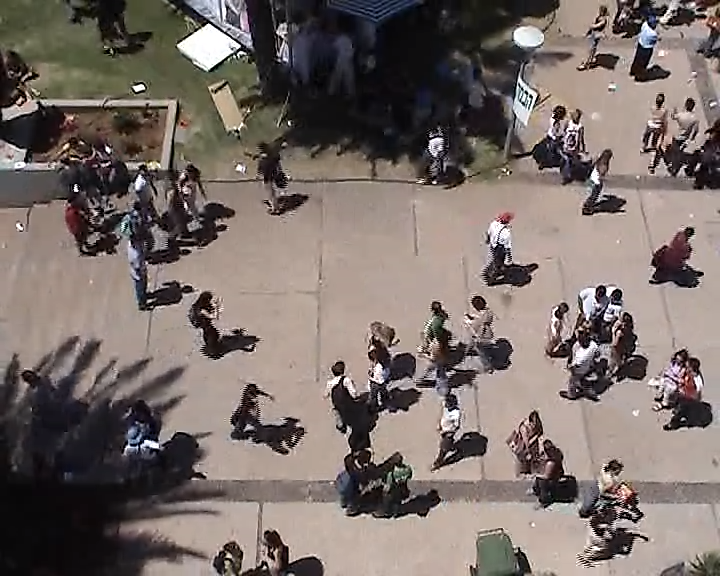}
    		\caption{Area of Campus dataset}
    		\label{fig:campus_area}
        \end{subfigure}
        \qquad
        \begin{subfigure}[b]{0.3\textwidth}
        	\includegraphics[width=\textwidth]{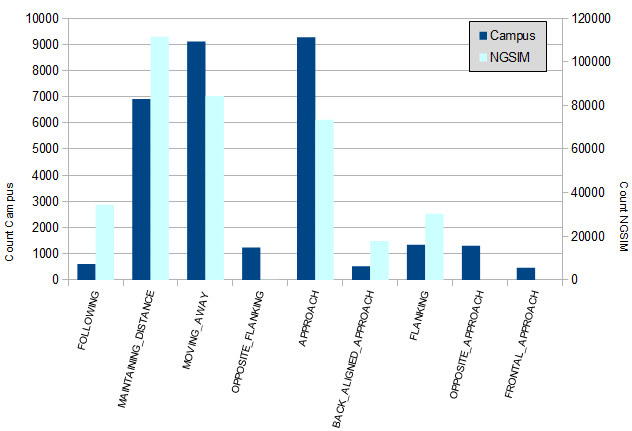}
    		\caption{Events found for the two datasets}
    		\label{fig:events_found}
        \end{subfigure}
        ~ 
        \label{fig:general_information}
        \caption{An overview on the considered datasets}
\end{figure*}

The first dataset we used is called NGSIM, and contains data about vehicles moving on 2100-feet segment of the US Highway 101 and covering a period of 15 minutes.
In Figure~\ref{fig:ngsim_area} we can see a photo of the area considered, that is substantially a straight road with an entering and an exit ramp. The locations of vehicles moving on the area have been sampled 10 times per second.

The second dataset represents a set of pedestrians moving in a small square of a university campus and covers a period of 3 minutes and 31 seconds.
In Figure~\ref{fig:campus_area} we show a video frame of the area under investigation. The locations of individuals are sampled 25 times per second.

All the software, datasets and additional information can be downloaded at: \url{http://kdd.isti.cnr.it/IPMiner}.

The two datasets involve very different environments, with different types of moving agents, yet we could use the same interaction functions and event templates for both. 
The values of the parameters for the interaction functions defined in Section~\ref{sec:ipa} have been tuned empirically through a series of preliminary experiments over each of the two datasets, and are reported in Table~\ref{table_values}.


\begin{table*}
\renewcommand{\arraystretch}{1.3} 
\caption{Parameters used for the instantiation of the IPA framework and for the algorithms for both datasets.} 
\label{table_values} 
\centering
\scalebox{0.75}{
\begin{tabular}{c || c || c || c || c || c || c || c || c || c || c  || c || c || c || c || c} 
\hline 
\bfseries dataset & \bfseries $r_{agent}$ & \bfseries $d_{search}$ & \bfseries $\epsilon_{move}$ & \bfseries $\epsilon_{dir}$ & \bfseries $\epsilon_{align}$ & \bfseries $\epsilon_{\parallel}$ & \bfseries $\epsilon_{dist}$ & \bfseries $\epsilon_{route}$ & \bfseries $\epsilon_{vel}$ & \bfseries $\epsilon_{lat}$ & \bfseries $t_{supp}$ & \bfseries $t_{min}$ & \bfseries $t_{e\!-\!supp}$ & \bfseries $min_{jc}$ & \bfseries $t_{win}$\\ 
\hline\hline 
NGSIM & 1.83m & 25m & 0.1m & 5$^{\circ}$ & 2m & 4$^{\circ}$ & 0.1m & 4$^{\circ}$ & 0.15 km/h & 3.25m & 0.8 & 20 $\frac{1}{10}$sec & 0.9 & 0.6 & 150 $\frac{1}{10}$sec\\ 
\hline 
Campus & 0.6m & 10m & 10$^{-4}$m & 13$^{\circ}$ & 1m & 13$^{\circ}$ & 0.01m & 13$^{\circ}$ & 0.005 m/s & 1.2m & 0.8 & 25 $\frac{1}{25}$sec & 0.9 & 0.6 & 300 $\frac{1}{25}$sec\\ 
\hline
\end{tabular}}
\end{table*}

\subsection{Analysis on NGSIM vehicle dataset}
The analysis with the IPA framework returned 343441 different event instances of duration at least one-tenth of second. The light (cyan) bars in Figure~\ref{fig:events_found} show how these instances are distributed within the different event categories.
We can see that {\em maintaining distance}, {\em moving away} and {\em approach} are the most frequent events, which seems reasonable for this specific context.
Also, no events involving opposite directions (e.g. {\em opposite approach}, etc.) were found, as could be expected, since the general direction of movements is the same for all cars as imposed by the road network.

On top of these events we launched the SIPM algorithm which extracted 98 static interaction patterns involving up to six agents per pattern. Among them, we report some examples in the top rows of Table~\ref{table_examples}.

\begin{figure*}[t]
        \centering
        \begin{subfigure}[b]{0.3\textwidth}
                \includegraphics[width=\textwidth]{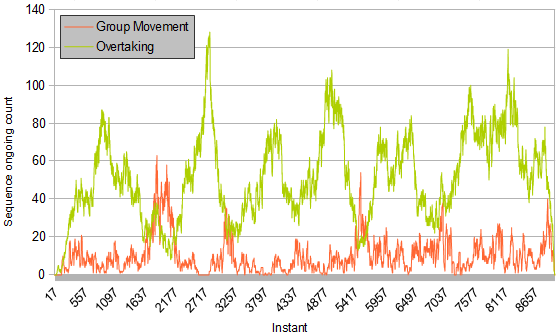}
                \caption{Ongoing events count}
    			\label{fig:ongoing_ngsim}
        \end{subfigure}%
        ~ 
        \begin{subfigure}[b]{0.3\textwidth}
        		\includegraphics[width=\textwidth]{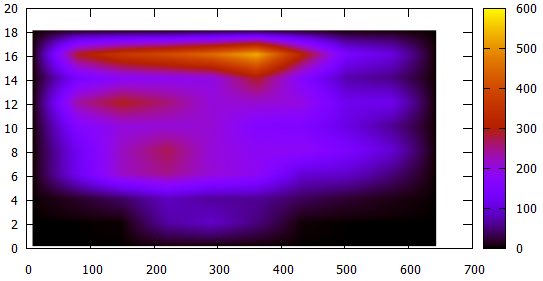}
    			\caption{Spatial density of {\em overtaking}}
    			\label{fig:spatial_ngsim_overtaking}
        \end{subfigure}
        ~ 
        \begin{subfigure}[b]{0.3\textwidth}
                \includegraphics[width=\textwidth]{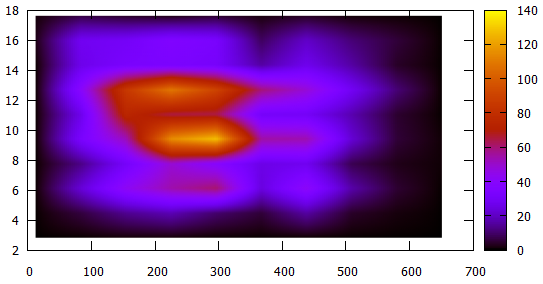}
    			\caption{Spatial density of {\em group movement}}
    			\label{fig:spatial_ngsim_group}
        \end{subfigure}
        \caption{(a) Temporal distribution of {\em overtaking} (green line) and {\em group movement} (red line). (b) and (c): spatial density distribution of the two patterns.}
\end{figure*}

Some of these patterns can be easily recognized.
For instance, in the first one (top left) an agent $A$ is following an agent $B$ and at the same time a third agent $C$ is in the middle, maintaining the same distance from both. 
This pattern represents a group moving together. 
However, other patterns are less intuitive, such as the fifth one (bottom left): two agents $D$ and $E$ are moving away from an agent $C$, which is approached by an agent $A$ which, in turn, is moving away from a fifth agent $B$. 

The static interaction patterns found were used as input for the EvIPM algorithm to find evolving patterns. By limiting the search to sequences of size up to 4 static pattern instances, the algorithm found 950352 sequence instances, grouped into 786 evolving patterns. 
The most frequent sequence is shown in the first row of table \ref{table_examples} (right column); this pattern has been seen over 6079 times and indicates an overtaking in which the flanking phase between the {\em approach} and the {\em moving away} was too fast (below 2 seconds) to be detected.
Actually, also the sequence $approach(A,B) \rightarrow flanking(A,B) \rightarrow moving\_away(A,B)$ (not reported here) was among the frequent patterns, yet with a much smaller frequency. 
The duration of the {\em flanking} phase can somehow characterize the overtakings, and the result above suggests that in this highway segment fast overtakings are more common that slow ones.
In figure \ref{fig:spatial_ngsim_overtaking} we report the spatial distribution of this sequence -- here the flow of the vehicles is from left to right, while the ramps are in the lower part -- which is most concentrated on the external lane of the road, as one can expect. 
In the density plot, for each sequence its static pattern instances are represented by the centers of their bounding boxs, and the sequence instance itself is represented by the average point of such centers.

The second row of Table~\ref{table_examples} shows a sequence that represents the evolution of a small group that is moving together.
Looking at its spatial distribution, represented in Figure~\ref{fig:spatial_ngsim_group}, we can see that, this time, it is mostly concentrated in the central lanes of the road and in proximity of the exit/input junction. 

\begin{figure}[h]
	\centering
    \includegraphics[scale=0.35]{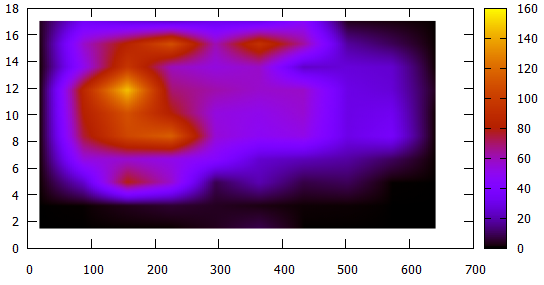}
    \caption{Density distribution for {\em complex overtaking}.}
    \label{fig:density_complex}
\end{figure}

Finally, in the last sequence shown in the Table~\ref{table_examples} for NGSIM, we can see an overtaking of an agent $A$ on an agent $B$ with other complex interactions in the middle. 
Correspondingly, the spatial distribution of this sequence, shown in Figure~\ref{fig:density_complex}, is very different from that of the simple overtaking, suggesting that they actually represent very different situations.

\subsection{Analysis on a campus crowd dataset}
For this dataset the analysis with the IPA framework has retrieved 30772 event instances, whose distribution across the event categories is shown in Figure~\ref{fig:events_found} by the dark (blue) bars.
We can see that the general distribution is concordant with the results obtained on the NGSIM dataset, yet here all the categories contain a significant number of instances, due to the fact that movement is free in this context.
The SIPM algorithm returned 48 static interaction patterns, containing up to five agents each. Some examples shown in the bottom rows of Table~\ref{table_examples}.

\begin{table*} 
\renewcommand{\arraystretch}{1.3} 
\caption{Sample interactions patterns and evolving patterns for the two datasets -- see Table~\ref{concrete_events} for a legend of symbols.} 
\label{table_examples} 
\centering 
\scalebox{0.8}{
\begin{tabular}{c || c : c || c} 
\hline 
DATASET & \multicolumn{2}{c}{INTERACTION PATTERNS} & EVOLVING INTERACTION PATTERNS \\ 
\hline\hline 
& \includegraphics[scale=0.35]{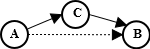} & \includegraphics[scale=0.35]{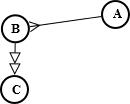} & \includegraphics[scale=0.35]{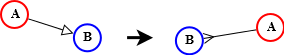}\\
\hdashline
NGSIM & \includegraphics[scale=0.35]{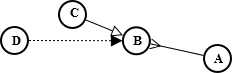} & \includegraphics[scale=0.35]{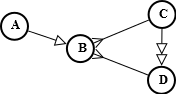} & \includegraphics[scale=0.35]{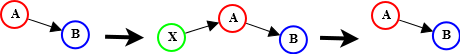}\\
\hdashline
  & \includegraphics[scale=0.35]{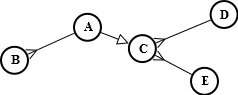} & \includegraphics[scale=0.35]{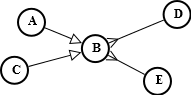} & \includegraphics[scale=0.35]{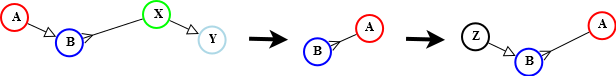}\\
\hline\hline 
Campus & \includegraphics[scale=0.35]{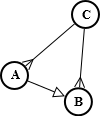} & \includegraphics[scale=0.35]{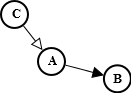} & \includegraphics[scale=0.35]{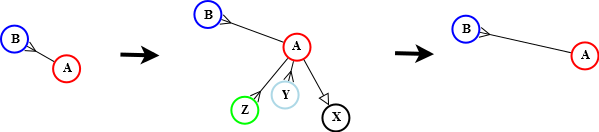}\\
\hdashline
  & \includegraphics[scale=0.35]{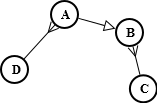} & \includegraphics[scale=0.35]{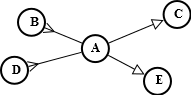} &  \includegraphics[scale=0.35]{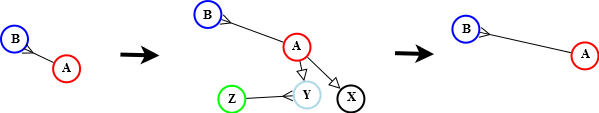}\\
\hline
\end{tabular}}
\end{table*}

\begin{figure}[h]
	\centering
    \includegraphics[scale=0.3]{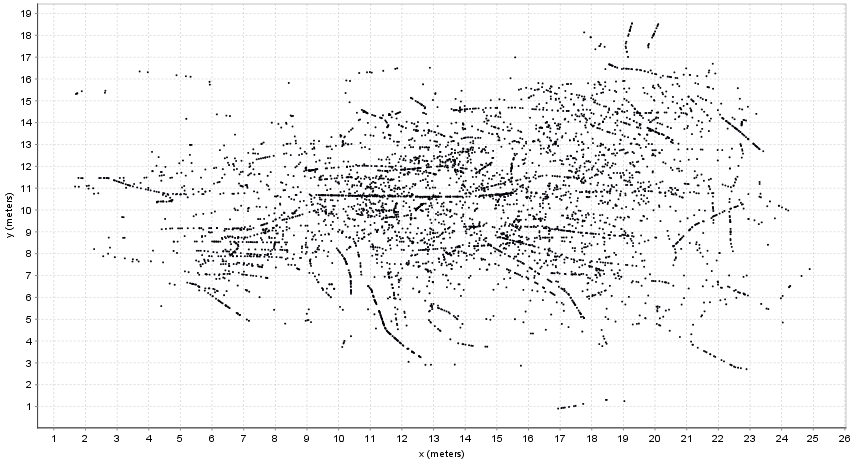}
    \caption{Locations of an evolving pattern on Campus.}
    \label{fig:spatial_campus}
\end{figure}

The most frequent static patterns found with a minimum temporal persistence of one second are very simple, and mainly involve {\em approach} and {\em moving away}. 
Applying the EvIPM algorithm, with a limit to sequences of at most 4 static pattern instances, we obtain 147839 sequences grouped into 171 evolving patterns.
The sequences found for this dataset are generally more difficult to interpret and to assign to clear labels (e.g. the overtaking) than those found on the NGSIM dataset, yet from a first analysis of the results they appear to be plausible interaction schemes for a place involving pedestrians.
An example is the last sequence reported in the Table~\ref{table_examples} (right column). 
Plotting the locations corresponding to the instances of this pattern (Figure~\ref{fig:spatial_campus}) we can recognize some groups of trajectories that move in similar directions, suggesting that there is some dependency between the kind of interaction that occurs the the direction of movement.

\subsection{SIPM and EvIPM performances}

\begin{figure*}[t]
        \centering
        \begin{subfigure}[b]{0.26\textwidth}
                \includegraphics[width=\textwidth]{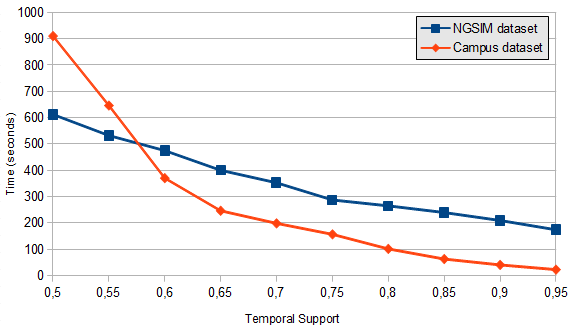}
                \caption{$t^{ngsim}_{min} = 2s$ , $t^{campus}_{min} = 1s$}
                \label{fig:ipm_time_varying_support}
        \end{subfigure}%
        ~ 
        \begin{subfigure}[b]{0.26\textwidth}
                \includegraphics[width=\textwidth]{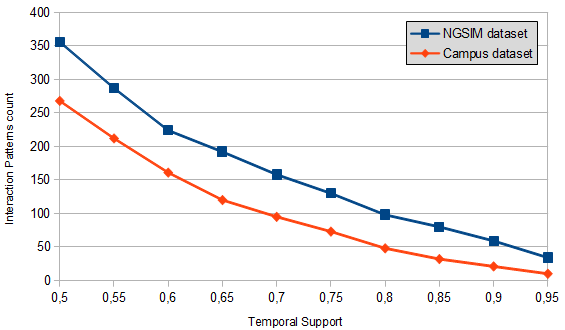}
                \caption{$t^{ngsim}_{min} = 2s$ , $t^{campus}_{min} = 1s$}
                \label{fig:tiger}
        \end{subfigure}
        ~ 
        \begin{subfigure}[b]{0.26\textwidth}
                \includegraphics[width=\textwidth]{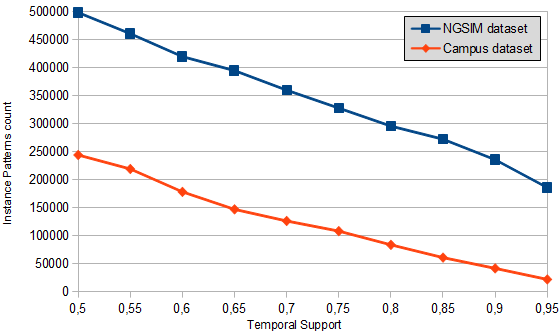}
                \caption{$t^{ngsim}_{min} = 2s$ , $t^{campus}_{min} = 1s$}
                \label{fig:mouse}
        \end{subfigure}
        \\
        \begin{subfigure}[b]{0.26\textwidth}
                \includegraphics[width=\textwidth]{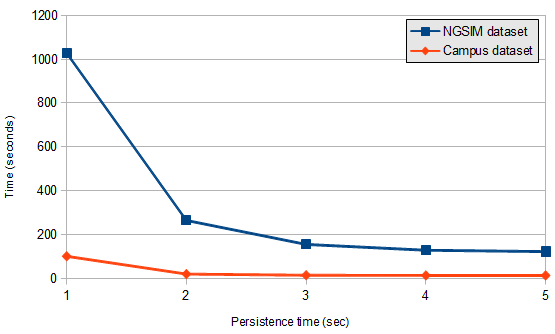}
                \caption{$t_{supp} = 0.8$ for both}
                \label{fig:gull}
        \end{subfigure}%
        ~ 
        \begin{subfigure}[b]{0.26\textwidth}
                \includegraphics[width=\textwidth]{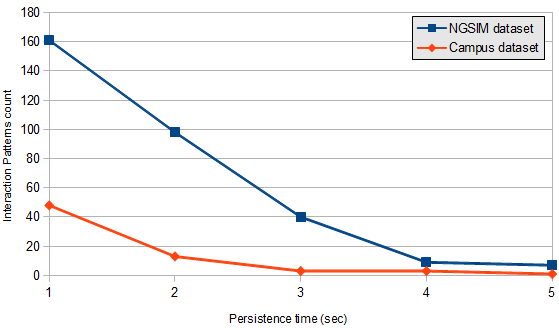}
                \caption{$t_{supp} = 0.8$ for both}
                \label{fig:tiger}
        \end{subfigure}
        ~ 
        \begin{subfigure}[b]{0.26\textwidth}
                \includegraphics[width=\textwidth]{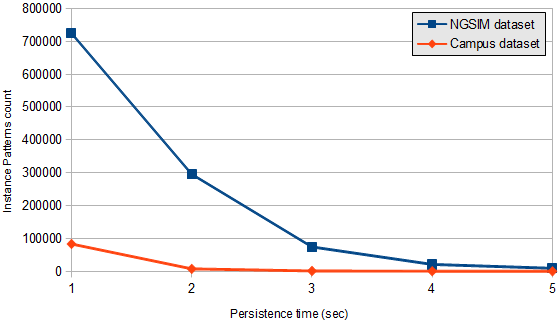}
                \caption{$t_{supp} = 0.8$ for both}
                \label{fig:mouse}
        \end{subfigure}
        \caption{SIPM performances w.r.t. time support $t_{supp}$ and persistence time $t_{min}$. The first column ((a) and (d)) shows run times, the second one ((b) and (e)) the number of static patterns, and the last one ((c) and (f)) the number of instances.}\label{fig:ipm_performances}
\end{figure*}

In this section we briefly report the performances test for the SIPM and the EvIPM algorithms. 
Figure~\ref{fig:ipm_performances} plots three metrics: execution times (plots on the left column), number of static patterns found (central column) and number of static pattern instances (right column).
On the first row of graphs, the metrics are shown in function of the minimum temporal support $t_\mathit{supp}$. On the second row, plots are in function of the minimum interval length (persistence) $t_{min}$ of the pattern instances.
In all plots, the blue lines refer to the NGSIM dataset, and the red ones to the Campus dataset.

We can see that the values of the two parameters considered have an important impact on performances, yet run times remain within acceptable limits in spite of the worst case computational complexity of the algorithm -- the most expensive computation required less than 20 minutes. 
The plots show also the critical importance of setting properly the value of $t_{min}$ (time persistence of instances), since larger values quickly reduce the set of output instances and patterns.

\begin{figure*}
        \centering
        \begin{subfigure}[b]{0.26\textwidth}
                \includegraphics[width=\textwidth]{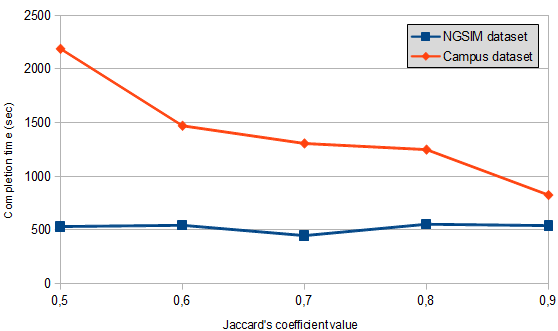}
                \caption{NGSIM: $t_{win} = 15s$, Campus: $t_{win} = 2s$, $t_{e\!-\!supp} = 0.9$ for both}
                \label{fig:ipm_time_varying_support}
        \end{subfigure}%
        ~ 
        \begin{subfigure}[b]{0.26\textwidth}
                \includegraphics[width=\textwidth]{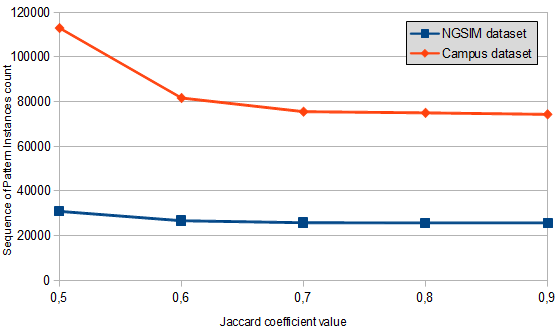}
                \caption{NGSIM: $t_{win} = 15s$, Campus: $t_{win} = 2s$, $t_{e\!-\!supp} = 0.9$ for both}
                \label{fig:tiger}
        \end{subfigure}
        ~ 
        \begin{subfigure}[b]{0.26\textwidth}
                \includegraphics[width=\textwidth]{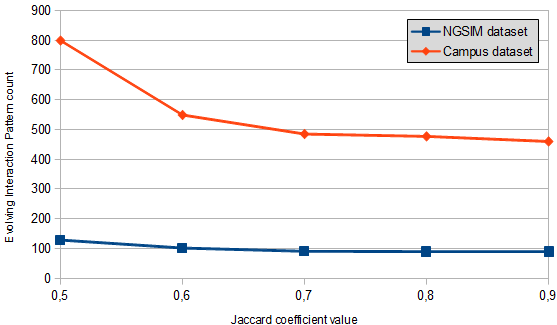}
                \caption{NGSIM: $t_{win} = 15s$, Campus: $t_{win} = 2s$, $t_{e\!-\!supp} = 0.9$ for both}
                \label{fig:mouse}
        \end{subfigure}
        \\
        \begin{subfigure}[b]{0.26\textwidth}
                \includegraphics[width=\textwidth]{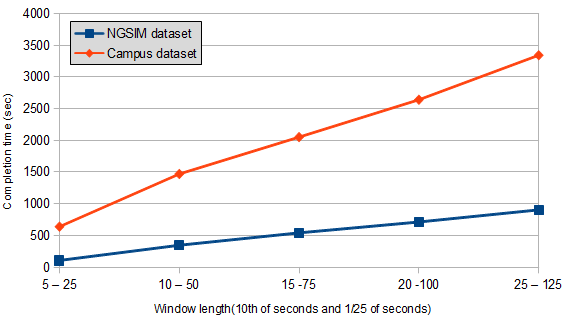}
                \caption{$min_{jc}=0.6$, $t_\mathit{e\!-\!supp}=0.8$}
                \label{fig:ipm_time_varying_support}
        \end{subfigure}%
        ~ 
        \begin{subfigure}[b]{0.26\textwidth}
                \includegraphics[width=\textwidth]{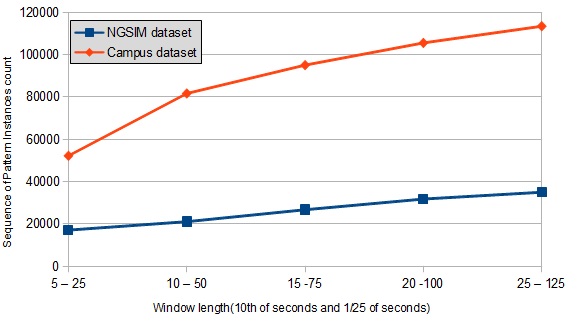}
                \caption{$min_{jc} = 0.6$, $t_{e\!-\!supp} = 0.8$}
                \label{fig:tiger}
        \end{subfigure}
        ~ 
        \begin{subfigure}[b]{0.26\textwidth}
                \includegraphics[width=\textwidth]{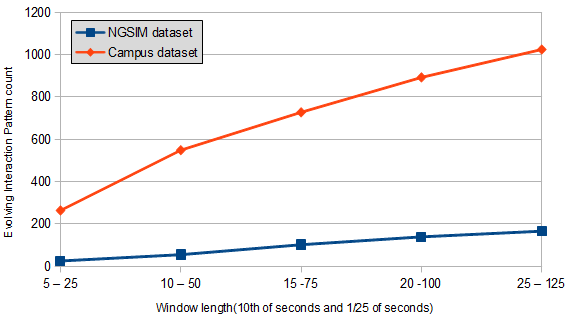}
                \caption{$min_{jc} = 0.6$, $t_{e\!-\!supp} = 0.8$}
                \label{fig:mouse}
        \end{subfigure}
        \caption{EvIPM performances w.r.t. Jaccard coefficient $min_{jc}$ and temporal window length $t_{win}$. The columns show, in the order: run times ((a),(d)), number of frequent evolving patterns ((b),(e)), and number of sequence instances found ((c),(f)).}\label{fig:seqipm_performances}
\end{figure*}

In Figure~\ref{fig:seqipm_performances} we plot the results of the tests performed on the EvIPM algorithm. 
Due to the higher complexity of this step, these experiments were performed on a 10\% sample of the datasets for both cases. 
In the first row we are varying the Jaccard coefficient threshold ($min_{jc}$), which appears to affect heavily the results over the Campus dataset, yet having very little impact over the NGSIM dataset. 
On the contrary, the temporal window threshold ($t_{win}$) has similar effects on both cases, resulting in larger result sets and run times for larger values of the parameter.
Surprisingly, the temporal support threshold $t_\mathit{e-supp}$ for the evolving interaction patterns (not plotted here for lack of space -- see~\cite{tesi_gabriele} for details) has a minor impact on performances.
That suggests that the pruning mechanism applied to sequence instances plays the major role in determining execution times, while the constraints on the temporal support of evolving patterns do not translate to significant pruning.

\section{Related works}
\label{sec:related_works}

The main inspiration for this paper comes from the {\bf simulation of crowds} field.
This topic is studied by a large body of literature aimed at better understanding and modelling the behaviour of masses of people in specific contexts, for instance during dangerous or panic situations~\cite{helbing_panic_nature, lakoba_modification_helbing}. 
This becomes very important to prevent those situations where uncoordinated movement can put at risk human lives. 
This kind of works generally assume that a crowd can be modelled similarly to physical systems (e.g., particles in isolated systems) guided by physical laws representing the interactions between the moving objects (particles, etc.).
A second approach is given by Agent Based Models~\cite{abm}, made of a set of agents that can take decisions individually by integrating physical models with a set of decision rules.
For instance, \cite{moving_agents_space_time} divides the aspects relative to the physical movement of the agent from the those related to path planning and interactions with other agents in their neighborhood, based on criteria such as the probability of collision and willingness to follow a group.
Similarly, yet on a different domain, works such as~\cite{ngsim} develop behavioural rules -- fitted to data -- that drive the lane selection process of cars on highways.
The framework proposed in this paper is quite complementary to these approaches, in that it tries to infer information about interactions directly from the data, rather than starting from the formulation of complete models that only later are (in some cases) compared with real observations.
We believe that our methods can positively contribute to this area, by helping in the process of reformulating or refining such models.


Another field related to our work is {\bf modelling complex systems}, which extends the scope of modelling beyond mobility. As an example, \cite{reality_mining} studies complex social systems through the information obtained from the cellular phones of a sample population, and try to infer social patterns in user activities, significant locations or habits of the users. From those data the authors tried also to infer relationships and communities for users.
Works like \cite{longhi2020} exploit the indirect interactions between drivers on the road as cues for estimating crash risk probabilities, by inferring various mobility-related indexes, especially inferred through trajectory mining methods \cite{geopkdd2010}
In this area, interaction is considered and also plays an important role. However, the user interactions considered are mainly in the form of explicit exchanges of information (phone calls, messages), with just a very simplified view on physical interactions, for instance reduced to detecting co-location of individuals.

From a more technical perspective, the analysis methods we proposed on top of simple interaction events involve temporal and graph-based information.
{\bf Temporal graph analysis}~\cite{Holme} is one important area that tries to analyze both components (time and network) together. 
Most solutions focus on simple properties of the (evolving) network, such as connectivity and distances, yet some work on {\em persistent subgraphs}~\cite{persistent_subgraphs} exists, which has similarities with our {\em static patterns}.
The main differences between them is that in our definition events and patterns are associated to time intervals, instead of unrelated sets of time instants, and such intervals can derive from complex and very flexible criteria (see {\em event templates} in Definition~\ref{def:event_templates}).
Moreover, the evolution of such patterns is an aspect that was not considered in~\cite{persistent_subgraphs}.
Other relevant approaches in literature consider only one of the two components at a time: {\bf temporal interval patterns} \cite{GQ2011,Pan+2009,Rigotti2006} look for recurrent arrangements of interval-based events, thus abstracting from the origin of the events themselves; on the other hand, {\bf frequent graph mining}~\cite{Jiang} focuses on the search of recurrent subgraphs in a single large input graph or in a database of (smaller) graphs.
Both lines of work, however, do not apply directly to our case, since they cannot capture our notion of (graph-based) pattern that persists and possibly evolves involving the same agents.

\section{Conclusion}\label{sec:conclusion}

In this paper we proposed a data-driven methodology and algorithms for the understanding of interactions between moving agents.
The general framework was also instantiated to a specific context, and then tested on two datasets of different nature.
The results of our experiments look promising, and confirm the ability of our approach to capture some well-known phenomena as well as some non-trivial ones.

The results obtained so far open a wide spectrum of opportunities for improving the methodology, as well as to move it closer to specific modelling problems. Among the various prospective lines of future work, we believe the following one to be the most interesting: 
(i) expand the experimentation both in width, i.e. including other scenarios, and in depth, i.e. extending the set of functions and event templates that instantiate the framework; 
(ii) extend the methodology by including the influence of the environment, such as obstacles, etc.; 
(iii) evaluate the applicability of the general framework beyond the mobility context, e.g. in finance applications, games, etc.;
(iv) finally, study the integration of our interaction patterns (or some refined variant) with existing simulation models, e.g. the models presented in ~\cite{helbing_panic_nature, lakoba_modification_helbing} include forms of mutual interaction among agents purely based on physical laws, which could be extended by considering also data-driven social components.

\begin{thebibliography}{1}

\bibitem{helbing_panic_nature} D. Helbing, I. Farkas, T. Vicsek, ``Simulating dynamical features of escape panic'', Nature, Volume 407, pp. 487-490, 2000

\bibitem{lakoba_modification_helbing} T. I. Lakoba, N. M. Finkelstein, ``Modifications of the Helbing-Molnar-Farkas-Vicsek social force model for pedestrian evolution'', Simulation, Volume 81, Issue 5, pp. 339-352, 2005

\bibitem{abm} E. Bonabeau, ``Agent-based modeling: Methods and techniques for simulating human systems'', Proceedings of the National Academy of Sciences of the United States of America, Volume 99, 2002

\bibitem{moving_agents_space_time} P. M. Torrens, ``Moving agent pedestrians through space and time'', Annals of the Association of American Geographers,Volume 102, Issue 1, pp. 35-66, 2012


\bibitem{reality_mining} N.Eagle, A. Pentland, ``Reality mining: sensing complex social systems'', Personal and Ubiquitous Computing, Volume 10, Issue 4, pp 255-268, 2006


%
%
%
%
%

\bibitem{tesi_gabriele} G. Galatolo. ``Data-driven exploration of mobility interaction-patterns'', Master Thesis at Dipartimento di Informatica, University of Pisa, 2014. Downloadable at: \url{http://kdd.isti.cnr.it/IPMiner}.

\bibitem{persistent_subgraphs}
M. Lahiri and T. Y. Berger-Wolf, ``Structure prediction in temporal networks using frequent subgraphs``, in IEEE Symposium on Computational Intelligence and Data Mining, pages 35--42, 2007.

\bibitem{GQ2011}
T.~Guyet and R.~Quiniou,
``Extracting temporal patterns from interval-based sequences'', In IJCAI'11, pages 1306--1311. AAAI Press, 2011.

\bibitem{Pan+2009}
P.~Papapetrou, G.~Kollios, S.~Sclaroff, and D.~Gunopulos,
``Mining frequent arrangements of temporal intervals'',
In {\em Knowledge and Information Systems}, 21(2):133--171, 2009.


\bibitem{Kostakos}
V. Kostakos, ``Temporal graphs'', In Physica A: Statistical Mechanics and its Applications, 388(6) , pp. 1007-1023, North-Holland, 2009.

\bibitem{Holme}
P. Holme, J. Saram\"{a}ki, Jari (Eds.),
``Temporal Networks'', Springer, 2013.

\bibitem{Jiang}
C. Jiang, F. Coenen and M. Zito, ``A survey of frequent subgraph mining algorithms'', in The Knowledge Engineering Review, v.28, pp 75-105.


\bibitem{longhi2020}
L. Longhi, M. Nanni, ``Car telematics big data analytics for insurance and innovative mobility services'', J. Ambient Intell Human Comput 11, 3989–3999 (2020). https://doi.org/10.1007/s12652-019-01632-4

\bibitem{geopkdd2010}
M. Nanni, R. Trasarti, C. Renso, F. Giannotti, and D. Pedreschi, ``Advanced knowledge discovery on movement data with the GeoPKDD system'', Proceedings of the 13th Int Conf on Extending Database Technology (EDBT 2010). ACM, USA, 693–696. https://doi.org/10.1145/1739041.1739129

\bibitem{ngsim} C. Choudhury, T. Toledo, M. Ben-Akiva, ``NGSIM freeway lane selection model'', Tech. Rep. FHWA-HOP-06-103 of the US FHA, 2004, http://ngsim-community.org

\bibitem{Rigotti2006}
M. Nanni, C. Rigotti, ``Extracting Trees of Quantitative Serial Episodes'', Lecture Notes in Computer Science, vol 4747. Springer, Berlin, Heidelberg, 2006. https://doi.org/10.1007/978-3-540-75549-4\_11

\end{thebibliography}

\end{document}